\documentclass[runningheads]{llncs}
\usepackage[T1]{fontenc}
\usepackage{graphicx}
\usepackage{caption} 
\usepackage{lscape} 
\usepackage{url}
\begin{document}
\title{Reasoning in Transformers -- Mitigating Spurious Correlations and Reasoning Shortcuts}
\titlerunning{Mitigating Spurious Correlations and Reasoning Shortcuts}
%
\author{Daniel Enström\inst{1}
\and Viktor Kjellberg\inst{1}
 \and Moa Johansson\inst{2} }
 \institute{
   University of Gothenburg, Gothenburg, Sweden. \\
	\email{\{gusensda, guskjevia\}@student.gu.se}
\and
   Chalmers University of Technology, Gothenburg, Sweden. \\
	\email{moa.johansson@chalmers.se}
}
\authorrunning{Enström, Kjellberg and Johansson}

%
%
%
\maketitle             
\begin{abstract}
Transformer language models are neural networks used for a wide variety of tasks concerning natural language, including some that also require logical reasoning. 
However, a transformer model may easily learn spurious patterns in the data, short-circuiting actual reasoning.
In this paper we investigate to what extent transformers can be trained to a) approximate reasoning in propositional logic while b) avoiding known reasoning shortcuts via spurious correlations in the training data. 
To do so, we use a dataset with known spurious correlation between truth and e.g. the number of rules in the problem. We augment the data with proofs, and train two models: a generative transformer, WP-BART, trained on problems and their whole proofs, and a neuro-symbolic model, SIP-BART, trained on individual proof steps and combining the generative transformer model BART with a symbolic proof checker. We find that SIP-BART succeeds in avoiding reasoning shortcuts, while WP-BART does not.
For SIP-BART, we then identify a few remaining reasoning errors, not previously described in the literature, arising from using a pre-trained language model. These are qualitatively analysed to create a taxonomy of four different types of additional pitfalls.
\end{abstract}

\section{Introduction}

Transformer models \cite{https://doi.org/10.48550/arxiv.1706.03762}, have successfully been applied to a range of tasks in natural language processing \cite{britz2017massive,radford2019language}, including those that requires the model to approximate reasoning \cite{saxton2019analysing}. 
It is generally acknowledged that
inducing large language models to reason step by step improves their results on reasoning tasks in natural language and when solving maths problems \cite{kojima2022large,polu2020gen,nye2021work,Rabe2020,wei2022}.  
We are interested in understanding how and why step-by-step reasoning helps transformers solve logic
problems. Does it help the model avoid learning spurious patterns and instead learn to better approximate actual reasoning?  

In this study, we consider problems in propositional logic, expressed in structured natural language. 
We are \emph{not} suggesting that transformer models are anywhere near specialised solvers on problems in propositional logic, and this is not the purpose of this study. Rather, we are interested in the reasoning capabilities (and pitfalls) of models processing \emph{primarily} natural language, but also able to do at least \emph{some} reasoning on problems expressed in English.
 Transformers have been reported to successfully having learnt to reason about the truth of these types of logic queries, essentially by classifying the query as True or False \cite{clark2021softReasoniners,talmor-leapofthought}.
 However, it is not always clear if this emergent functionality corresponds to a model really having learnt deductive reasoning, or if it is picking up on some other pattern in the data. Zhang et al. \cite{zhang2022paradox}, have experimentally shown that transformers are susceptible to learning spurious patterns inherent to reasoning problems, such as the number of rules in the problem description. With more available rules, chances are bigger that at least some of them will apply to the query and help prove it. So, it is not a bad heuristic for \emph{guessing} if a query is true, but it does not constitute logical reasoning as we know it. They demonstrate this using a BERT classifier \cite{devlin-etal-2019-bert}, trained on a dataset called SimpleLogic, consisting of propositional reasoning problems expressed in natural language and divided into subsets with different (spurious) statistical patterns correlating with the truth of the query. The BERT model fails to generalise between subsets and they conclude that it is difficult to ensure that transformers really do learn to reason: there are countless potential spurious patterns present in reasoning problems. 
This lack of reliable deductive reasoning capabilities in large language models has been argued to be a significant obstacle for their further development \cite{https://doi.org/10.48550/arxiv.2206.10498,https://doi.org/10.48550/arxiv.2102.01017}.

 We experiment with ways of addressing this problem: reasoning ought to be robust also over data with spurious correlations between problem properties and truth values. First, we augment the training data in SimpleLogic to include not only a True/False label but also the proof steps generated by a standard forward chaining algorithm. This leads to our first research question: 
 \textit{(1) To what extent can a generative transformer model avoid learning spurious correlation if trained to also generate short proofs?}
 
 Next, we also investigate a neuro-symbolic architechture more closely coupling the transformer with a symbolic checker. Here, instead of training on the whole proof at once, the generative neural model is trained to  produce intermediate inference steps, one at the time, which are iterated via a rule-based symbolic system to update the proof state \cite{susskind2021neurosymbolic,tafjord2021proofwriter,creswell2022selectioninference}.  This leads to our second research question: \textit{(2) Can a generative transformer model fully avoid learning spurious correlations if train to generate only one proof step at the time?} 
 
 Previous work has showed that training on step-by-step reasoning should improve accuracy on benchmarks, but no investigation has been made into the errors still made by the model. For instance, it may also be the case that the model correctly conclude a query being true/false, but still produce some faulty proof-steps along the way. This leads to our third research question: \textit{(3) What are the sources of other consistency errors and how can we avoid them?}

\section{Method}
We first introduce our augmented dataset SimpleLogicPS (SimpleLogic with Proof Steps), which extends the original SimpleLogic \cite{zhang2022paradox} with inference steps. We then describe our models, one generative transformer trained to produce whole proofs (WP-BART), and one neuro-symbolic model which combines a generative model for selecting inferences, with a symbolic module for keeping track of the proof state (SIP-BART). Finally, we describe our evaluation strategy.  Additional details regarding the models can be found in the MSc thesis of the first two authors \cite{MScthesis}, and the code is available online.\footnote{\url{https://anonymous.4open.science/r/paradox_alvis-FF17/}}

\subsection{Data}
The original SimpleLogic dataset \cite{zhang2022paradox}, consists of 840 000 simple inference problems in propositional logic. Each problem consists of a set of \emph{rules}, expressed as Horn clauses with 1-3 premises, a set of \emph{facts} (literals that are True) and a query literal for which a truth value is to be determined. We operate under a closed-world assumption, meaning that any query for which truth cannot be derived from the existing rules and facts, is assumed to be False.  The \emph{depth} of a problem refers to the minimum number of rule applications required to prove the query. For False queries, the depth refers to the maximum depth of the shallowest failing branch for that problem. See Appendix \ref{AppA} for an illustration. 
 
 SimpleLogic is divided into three sub-sets, equal in size, with different statistical features. Each subset is divided with a 80-10-10 split for training, validation and testing.
    \begin{description}   
     \item [Rule-Priority (RP)] problems were generated by first randomly generating rules, facts, and a query and then computing its label by forward chaining. 
     \item [Label-Priority (LP)] problems were generated in the opposite way as RP, i.e. first determining a query and its truth value and then randomly sampling rules and facts consistent with this.
  \end{description}
  For both RP and LP, Zhang et al. shows that the number of rules and facts, as well as the branching factor of proofs are spurious statistical features correlated with the truth value of the query \cite{zhang2022paradox}. They thus also introduce a third subset:
   \begin{description}   
     \item[Balanced Rule-Priority (RP\_b)] is similar to RP but the correlation between the number of rules and the boolean labels of queries has been removed.
 \end{description}

In the original SimpleLogic dataset, the target label is just the boolean truth value of the query. To give the generative model increased guidance during training the dataset is augmented with proofs. We have thus extended all problems with inference steps to allow the transformer to learn to identify the next inference rule to apply, rather than classifying the whole query as True or False at once. 
The proof steps in SimpleLogicPS were generated using forward chaining from known facts until the query was found (True queries), or until the search space was exhausted (False queries). 

\begin{figure}[ht]
    \centering
    \includegraphics[scale=0.1]{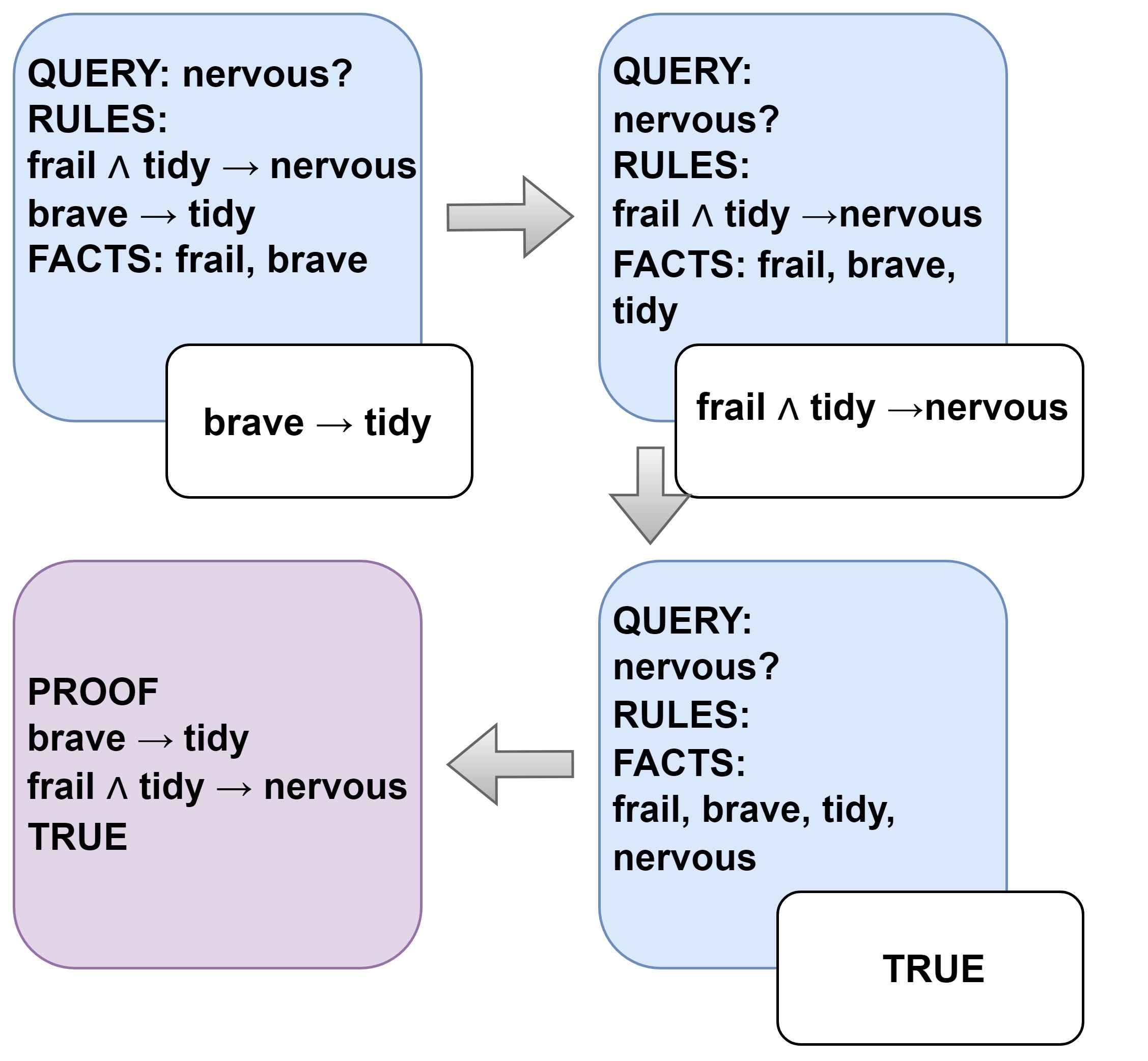}
    \caption{Example of how the generation procedure works. The three blue boxes represent three instances of input strings in SimpleLogicPS and the white boxes represent their respective output strings. The neural model thus regard each step of the proof as an isolated problem to be solved, since each step is a separate training instance. The complete proof string created by the model when inference is finished for a given problem is represented by the purple box.}
    \label{fig:ex_gen_proc}
\end{figure}

For the neuro-symbolic model, which we train on single proof steps, rather than whole proofs, so each original problem was then divided into individual proof steps. Each training data point thus consist of an input problem with an applicable rule as desired output (or True or False if it represented the final step). This resulted in a total of 3 986 165 proof step training instances. Three such example instances along with a complete proof string can be found in figure \ref{fig:ex_gen_proc}.

\subsection{Model Design and Training}

 We experiment with two different models trained also on proof information. Both are based on a pre-trained generative transformer model called BART \cite{lewis2019bart}, which is similar in size to the BERT classifier used in \cite{zhang2022paradox}. BART can also generate short text outputs, in our case a proof or proof step, not just a true/false answer.  For comparison, we also replicate the experiment by \cite{zhang2022paradox}: training BERT based binary classifiers on each subset of SimpleLogic to produce a True/False label for each problem. 
 \subsubsection{Whole-Proof BART (WP-BART).} The WP-BART model is trained to generate the proof for the input problem as a string, ending in TRUE or FALSE. It is trained on SimpleLogic proofs generated by a simple backward chaining algorithm.

\subsubsection{Symbolic Iterative Proof-BART (SIP-BART).}
 
\begin{figure}[ht]
    \centering
    \includegraphics[scale=0.09]{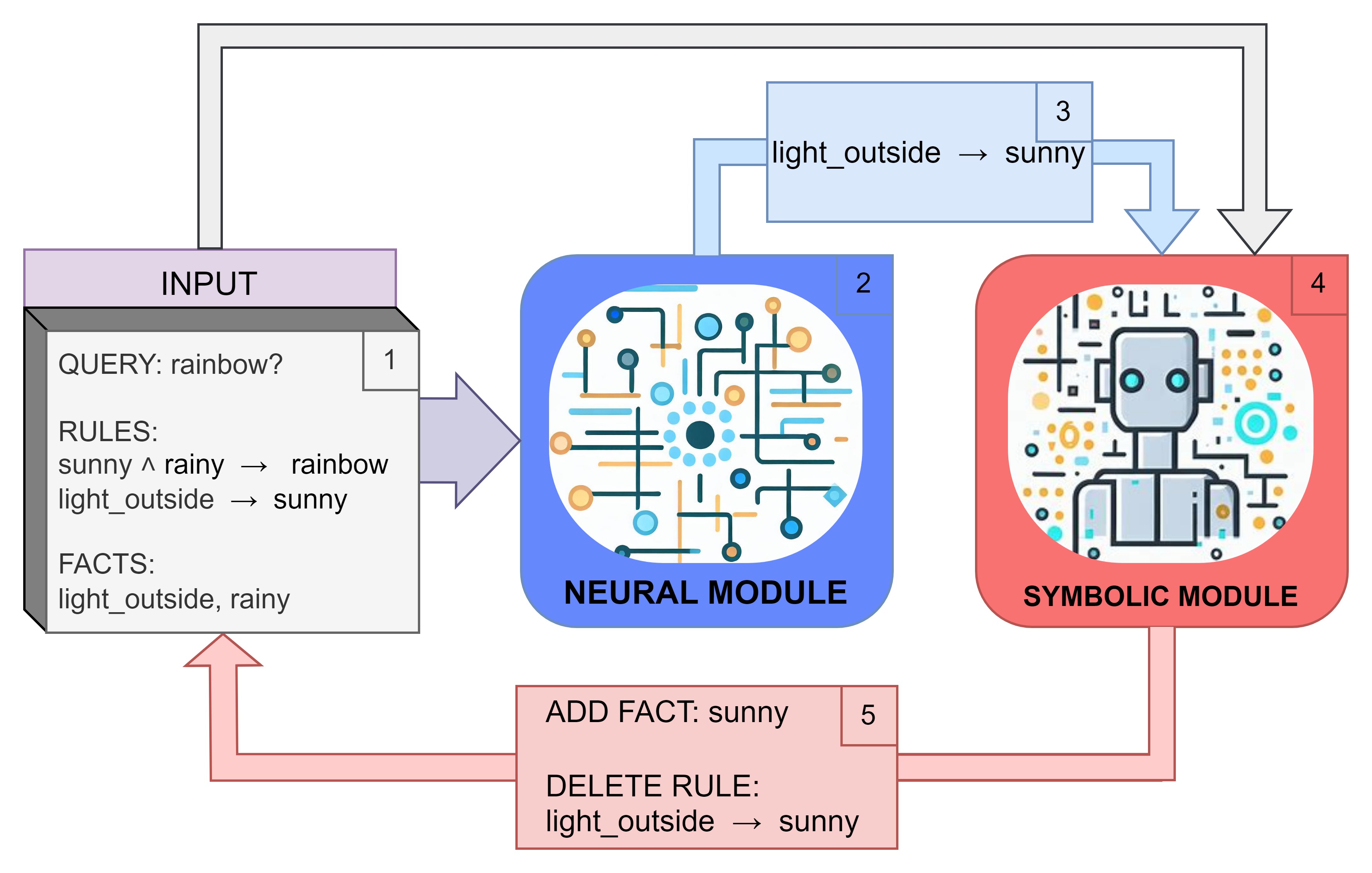}
    \caption{An overview of the SIP-BART architecture using an example of data-flow during inference.}
    \label{fig:sip-bart}
\end{figure}

The SIP-BART model combines the neural BART-component with a symbolic module responsible for updating the current state of the proof, see Figure \ref{fig:sip-bart}. 

 The neural module is a pre-trained BART model which is fine-tuned to generate the next applicable rule for SimpleLogicPS-problems as an output, a boolean string True when the proof is judged to be completed, or False when the search space is judged to be exhausted and no more rules are applicable. The symbolic module is responsible for applying the rule produced by the neural module, and produce a new textual representation for the next proof step, which serves as the next input to the neural model. Should the neural model produce a suggestion of a rule which is in fact \emph{not} applicable, the symbolic model records this faulty step, but does not update the state, and asks the neural model to "try again". This is however very rare, most generated steps are valid (see analysis of consistency in \S \ref{sec:consistency}).
 This iteration continues until the neural model outputs True or False, or the number of iterations reaches a maximum threshold to avoid potential non-termination\footnote{This was set to 100 iterations, however in practice a much lower threshold could have been used, as non-applicable suggestions turned out to be rare.}. The symbolic module then terminates the generation and outputs the final sequence of inference steps. If the system fails to return a chain of inferences ending in True or False the result is treated as a False Positive or False Negative depending on the ground truth label.
 
  \subsubsection{Training.} We train one WP-BART/SIP-BART model for each subset of SimpleLogic: LP, RP and RP\_b. We denote the differently trained models by WP-BART(LP)/SIP-BART(LP) etc. Each model was then evaluated on all subsets to be able to analyse to what extent the models learned spurious correlations from the data. The models were all trained for one epoch (as additional epochs did not improve results) on four NVIDIA Tesla A40 GPUs. 
  Refer to the Appendix \ref{AppB} for detailed information about training parameters and hardware. 

\subsection{Evaluation}
\label{sec:eval}
Evaluation was performed in two stages. First, addressing research questions  1 and 2, we evaluate the correctness of the truth-value of the query. Here, we compare the accuracy of the BERT replication with the final labels produced by WP-BART and SIP-BART (disregarding the intermediate details of the proof).

Secondly, to address our third research question, we analyse the sequence of inferences produced of our best performing architecture above. Here, the soundness and completeness of the inference steps were evaluated to check its consistency. A proof with no validity issues or extra faulty steps and no missing inference steps is regarded as consistent.  We analyse the errors in inconsistent proofs and identify four situations which give rise to inconsistencies:

\begin{description}
    \item[Non-existing Rule (NonExR).] Here, the neural model have suggested a rule which is not present in the problem description. 
    \item[Inapplicable Rule (InappR).] The neural model have suggested a rule which is present, but not applicable, as its premises are not satisfied (the relevant facts are not present in the problem description).
    \item [Spurious Match (SpMatch).] Here, the neural model have produced the final output \emph{True} to conclude the proof prematurely (the query is in fact not proved to be True yet - a spurious match with some other fact).
    \item[Unexhausted Search Space (UnexhS).] The neural model have prematurely produced the final output \emph{False}, while there are still rules available that would be applicable (the search space is not yet fully explored). 
\end{description}
 
 \section{Results}

\subsection{Accuracy of Truth Values}
We first assess the accuracy of the final labels (True or False) assigned to the queries of each problem in the SimpleLogic test sets.
The results of the replication experiment of \cite{zhang2022paradox} are shown Table \ref{tab:ex1_results}. We observe the same pattern in the data as is described in their article\footnote{As expected, despite using the same hyperparameter and seeds, we do not get exactly the same results. This is a know issue with neural network libraries and due to e.g. differences in hardware platforms used \cite{pytorch-repr}.} - a BERT classifier trained on specific distributions has a hard time generalizing to other distributions due to the model learning statistical features rather than approximating reasoning.
The model trained on RP and RP\_b generalised better to LP (total accuracy of 86.7 \% and  87.5\%  respectively) than LP to RP and RP\_b (total accuracy of 75.0 \% and 72.5\% respectively). 
Training on RP\_b result in slightly better performance than on RP. The accuracy also generally drops with increased proof-depth.
\begin{table*}[htbp]
\begin{tabular}{|*{10}{p{11mm}|}}
\hline
TRAIN   & TEST  & \textbf{0}         & \textbf{1}         & \textbf{2}         & \textbf{3}         & \textbf{4}         & \textbf{5}         & \textbf{6}     & \textbf{TOT}\\ \hline
LP & LP & 99.8 & 99.8 & 99.8 & 99.6 & 98.8 & 97.2 & 95.4 & \textbf{98.6} \\ \hline
LP & RP & 97.4 & 92.5 & 64.5 & 60.2 & 67.6 & 72.6 & 69.9 & \textbf{75.0} \\ \hline
LP & RP\_b & 97.7 & 93.3 & 60.2 & 56.7 & 63.9 & 68.7 & 68.5 & \textbf{72.7} \\ \hline
RP & LP  & 99.9      & 99.9      & 99.0      & 94.3      & 83.8      & 65.6      & 50.0    & \textbf{84.7}  \\ \hline
RP  & RP        & 99.8      & 100.0     & 99.4      & 98.9      & 98.6      & 96.9      & 95.9    & \textbf{98.5}  \\ \hline
RP & RP\_b & 99.2 & 99.2 & 98.6 & 98.0 & 96.6 & 93.9 & 89.1 & \textbf{96.4} \\ \hline
RP\_b & LP         & 99.7 & 99.4 & 99.3 & 96.4 & 87.6 & 72.6 & 57.2 &  \textbf{87.5}   \\ \hline
RP\_b & RP         & 99.8 & 99.9 & 99.5 & 98.9 & 98.6 & 97.9 & 96.9 & \textbf{98.8}    \\ \hline
RP\_b & RP\_b     & 99.6 & 99.5 & 99.0 & 98.4 & 98.0 & 96.7 & 94.1 &  \textbf{97.9} \\ \hline
\end{tabular}
\centering
\caption{Accuracy from the BERT-classifier replication of \cite{zhang2022paradox} trained on the different distributions. The integers in the heading refer to the depth of the ground-truth proof. TOT is the average accuracy across all proof depths. There are roughly the same number of problems with proofs of each depth. }
\label{tab:ex1_results}
\end{table*}

\subsubsection{Accuracy of WP-BART.}
Table \ref{tab:wp-bart} shows accuracy for the WP-BART model. It does not significantly improve overall compared to the baseline BART classifier. While WP-BART(LP) appears to generalise slightly better, the other variants perform worse. However, we also noticed that with increasing depth, the proportion of false queries in the data increase. The WP-BART(LP) model appears to have learned this as a reasoning shortcut, and over-predict the conclusion False, which can be seen in as an increased rate of False Negatives with proof depth, see Table \ref{tab:conf_mat_whole_LP_RP} in Appendix \ref{AppC}. We also note that the WP-BART(RP) model (trained on data where the number of rules correlated with truth) does not generalise as well as the model WP-BART(RP\_b) (trained on balanced data), suggesting that WP-BART is still prone to learn reasoning shortcut correlating the number of rules and truth.
\begin{table*}[htbp]
    \begin{tabular}{|*{10}{p{11mm}|}}
    \hline
    TRAIN   & TEST  & \textbf{0}         & \textbf{1}         & \textbf{2}         & \textbf{3}         & \textbf{4}         & \textbf{5}         & \textbf{6}     & \textbf{TOT}\\ \hline
    LP & LP & 100. & 100. & 92.6 & 90.2 & 89.8 & 91.2 & 93.3 & \textbf{93.9} \\ \hline
    LP & RP & 100. & 99.9 & 83.3 & 65.5 & 67.3 & 72.0 & 76.5 & \textbf{80.6} \\ \hline
    LP & RP\_b & 100. & 99.9 & 82.1 & 64.9 & 66.3 & 74.0 & 83.0 & \textbf{81.4} \\ \hline
    
    RP & LP & 84.7 & 85.4 & 79.2 & 73.9 & 71.6 & 68.5 & 63.4 & \textbf{75.2} \\ \hline
    RP & RP & 84.3 & 88.5 & 87.9 & 87.6 & 85.0 & 79.8 & 78.1 & \textbf{84.5} \\ \hline
    RP & RP\_b & 87.7 & 88.3 & 88.8 & 87.7 & 85.4 & 80.7 & 79.7 & \textbf{85.5} \\ \hline
    
    RP\_b & LP & 84.1 & 94.3 & 89.6 & 85.1 & 80.9 & 76.6 & 71.6 & \textbf{83.2} \\ \hline
    RP\_b & RP & 84.0 & 94.2  & 94.3 & 92.2 & 89.3 & 84.8 & 82.2 & \textbf{88.7} \\ \hline
    RP\_b & RP\_b & 87.2 & 93.6 & 94.1 & 93.3 & 88.6 & 86.8 & 85.7 & \textbf{89.9} \\ \hline
    \end{tabular}
    \centering
    \caption{Accuracy for the WP-BART models trained on the different distributions measured in percentage of correct predictions. The integers refer to the depth of the ground-truth proof. TOT is the overall average accuracy across proofs of all depths. There are roughly the same number of inference-problems for each depth.}
    \label{tab:wp-bart}
 \end{table*}

\subsubsection{Accuracy of SIP-BART.}
 The accuracy scores from the three SIP-BART models tested on all subsets are summarised in Table \ref{tab:LPgenstep}. There is a clear improvement in accuracy when compared to the BERT classifier in Table \ref{tab:ex1_results}. Additionally, the SIP-BART models achieved an almost perfect accuracy on all three testing sets, regardless of training set, with minimal or no drop as proof-depth increase. 
The accuracy for SIP-BART(RP) and SIP-BART(RP\_b) are equal, and slightly better than SIP-BART(LP) which performs worse on proofs of increased depth. Furthermore, no substantial over-prediction of conclusion False can be seen with the increased proof depths for SIP-BART(LP), see Table \ref{tab:conf_mat_step_LP_RP} in Appendix \ref{AppD}.
This indicates that, unlike the BERT classifier and WP-BART, SIP-BART has to a much larger extent \emph{avoided learning the spurious patterns} present such as the number of rules. The SIP-BART architecture is still not completely invariable to changes in distribution, as seen by the slightly lower accuracy for SIP-BART(LP) tested on RP and RP\_b. Though, it must be said this is only a minor issue - the accuracy of the different SIP-BART models are all above 99.8\% in total.  

 \begin{table*}[htbp]
    \begin{tabular}{|*{10}{p{11mm}|}}
    \hline
    TRAIN   & TEST  & \textbf{0}         & \textbf{1}         & \textbf{2}         & \textbf{3}         & \textbf{4}         & \textbf{5}         & \textbf{6}     & \textbf{TOT}\\ \hline
    LP & LP & 99.94 & 100. & 99.97 & 100. & 100. & 100. & 100. & \textbf{99.98} \\ \hline
    LP & RP & 99.97 & 99.97 & 100. & 99.92 & 99.90 & 99.872 & 99.49 & \textbf{99.87} \\ \hline
    LP & RP\_b & 100. & 99.97 & 99.97 & 99.89 & 99.94 & 99.74 & 99.25 & \textbf{99.81} \\ \hline
    
    RP & LP & 99.97 & 100. & 100. & 100. & 100. & 100. & 100. & \textbf{99.99} \\ \hline
    RP & RP & 100. & 100. & 100. & 100. & 100. & 100. & 99.97 & \textbf{99.99} \\ \hline
    RP & RP\_b & 100. & 100. & 100. & 100. & 100. & 100. & 100. & \textbf{100.} \\ \hline
    
    RP\_b & LP & 99.97 & 100. & 99.97 & 100. & 100. & 100. & 100. & \textbf{99.99} \\ \hline
    RP\_b & RP & 99.94 & 100. & 100. & 100. & 99.97 & 99.97 & 99.97 & \textbf{99.98} \\ \hline
    RP\_b & RP\_b & 100. & 100. & 100. & 100. & 100. & 99.97 & 100. & \textbf{99.99} \\ \hline
    \end{tabular}
    \centering
    \caption{Accuracy for the SIP-BART models trained on the different distributions measured in percentage of correct predictions. The integers refer to the depth of the ground-truth proof. TOT is the overall average accuracy across proofs of all depths. There are roughly the same number of inference-problems for each depth. 100. means exactly 100.. Values very close to 100. have been floored to 99.99.}
    \label{tab:LPgenstep}
 \end{table*}


  In summary, this reinforces the result by \cite{tafjord2021proofwriter}: a stepwise generation of the proof improves the models ability to solve logic based problems expressed in natural language. In addition, we also show that the \emph{reason} for this improvement is that the models are less likely to internalise spurious correlations that exist in the data. But even if the models are able to achieve a high accuracy they are not able to solve all problems, which indicates that there are still flaws in their approximation of reasoning. It is therefore necessary to further analyse what type of errors that are committed and the reasons for them. We proceed by doing so for our best performing model, SIP-BART. 

\subsection{Consistency of SIP-BART}
\label{sec:consistency}
 Although the accuracy tells us how well a given model is able to determine the truth-value of the query in each problem, it does not take into account if each of the inference steps themselves are sound and complete (i.e., consistent). 
  Even though the consistency scores where high, with all SIP-BART models achieving above 99\% fully consistent proofs, there are four types of reoccurring errors, as described in \S \ref{sec:eval}. The frequencies of these are shown in Table \ref{tab:ConsistencyErrors}.
Again, SIP-BART(LP) gets a somewhat lower score that compared to the other two models, with the lowest consistency of the series of inference steps being 99.11\% when tested on RP\_b.  The following sections will be dedicated to analysing each of these types and highlighting some of the reasons for why they occur. 
 \begin{table*}[htbp]
    \begin{tabular}{|p{10mm}|p{10mm}|p{15mm}|p{15mm}|p{15mm}|p{15mm}|p{15mm}|p{20mm}|p{20mm}|p{20mm}|}
    \hline
    \textbf{Train} & \textbf{Test} & \textbf{NonExR} & \textbf{InappR} & \textbf{SpMatch} & \textbf{UnexhS} & \textbf{Error Rate} & \textbf{Tot. Consistency} \\ \hline
    LP & LP & 0.036 & 0. & 0.007 & 0.004 & 0.046 & 99.954 \\ \hline
    LP & RP & 0.661 & 0. & 0.004 & 0.021 & 0.686 & 99.314 \\ \hline
    LP & RP\_b & 0.868 & 0. & 0. & 0.021 & 0.889 & 99.111 \\ \hline
    RP & LP & 0.018 & 0.004 & 0.004 & 0. & 0.025 & 99.975 \\ \hline
    RP & RP & 0.007 & 0.018 & 0. & 0.004 & 0.029 & 99.971 \\ \hline
    RP & RP\_b & 0.021 & 0.014 & 0.004 & 0. & 0.039 & 99.961 \\ \hline
    RP\_b & LP & 0.011 & 0. & 0.004 & 0.004 & 0.018 & 99.982 \\ \hline
    RP\_b & RP & 0.032 & 0.011 & 0.007 & 0.011 & 0.061 & 99.939 \\ \hline
    RP\_b & RP\_b & 0.025 & 0.007 & 0.004 & 0.004 & 0.039 & 99.961 \\ \hline
    \end{tabular}  
    \centering
    \caption{Frequencies of different types of consistency-errors in percent for the different training and testing sets. The 0.000 values are close to zero but not exactly zero. The 0. values are exactly zero.}
    \label{tab:ConsistencyErrors}
 \end{table*}

\subsubsection{Non-existing Rule}
\begin{figure}[htbp]
    \centering
    \includegraphics[width=8.5cm]{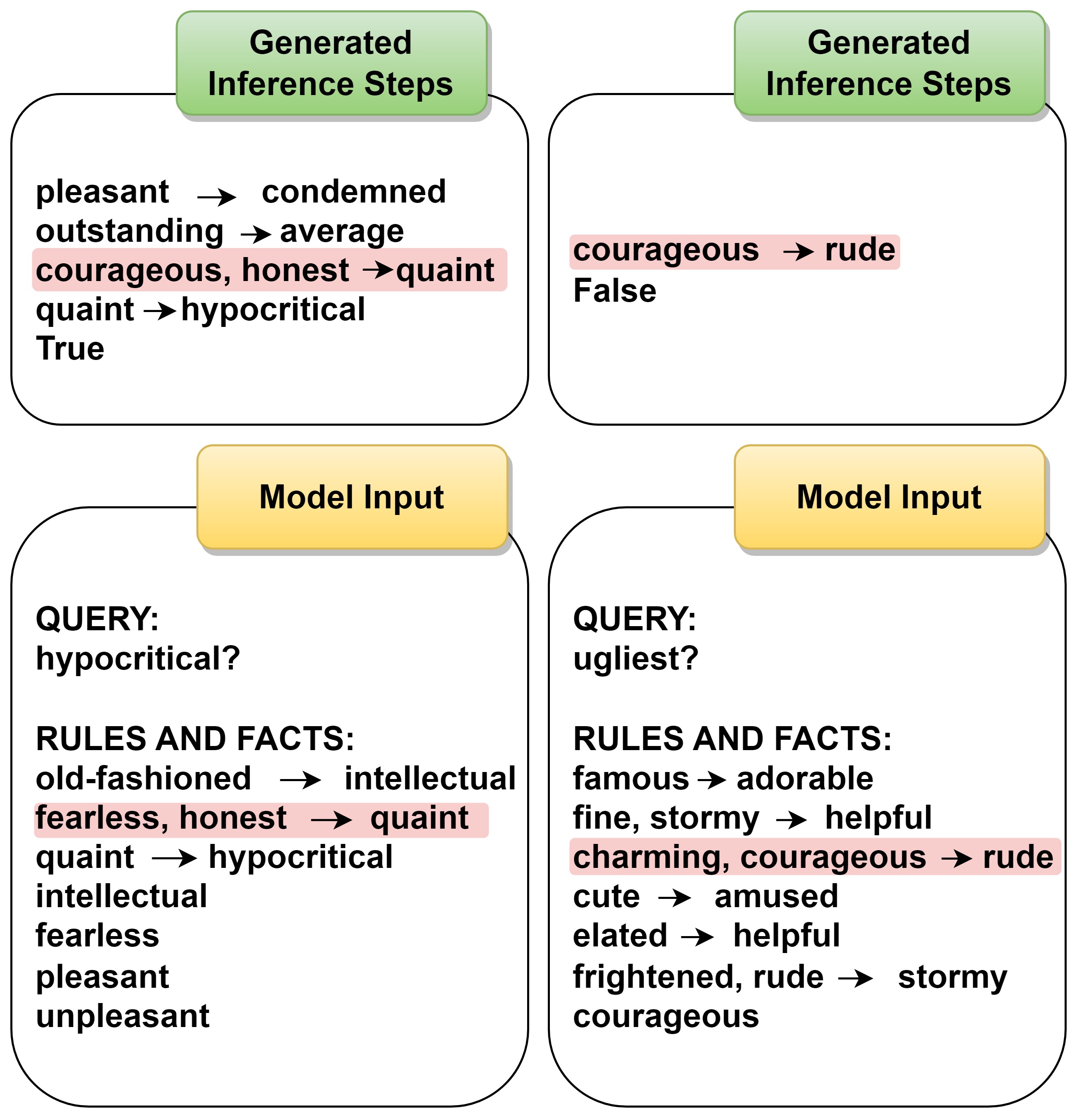}
    \caption{Two examples of where the Non-existing Rule error has occurred. On the left, the generated rule include a synonym word "courageous" instead of "fearless". The second example (right) the model generated part of the rule but missed the premise "charming".}
    \label{fig:NonexistingRule}
\end{figure}

 Across most of the SIP-BART models, the most common error was Non-existing Rule, which occurred in between 0.01\% - 0.87\% of proofs, depending on the test set. Based on qualitative analysis, it could be observed that the most common reason for this error was that the model had generated part of a existing rule but not the complete rule. The generated step often included the latter part of an existing rule but had missed one or more of the premises, see the right-side example in Figure \ref{fig:NonexistingRule}. Another error was that the model generated synonyms to the word in a rule instead of the word that existed in the original rule (Figure \ref{fig:NonexistingRule}, left), where the generated rule includes the word "courageous" instead of "fearless". Interestingly, most of the steps in the inference procedure that include non-existing rules are still almost correct (we can view them as near misses). In most cases, the model still provides a correct final classification of the problem, even when out-of-distribution.

\subsubsection{Inapplicable Rule}
\begin{figure}[htbp]
    \centering
    \includegraphics[width=8.5cm]{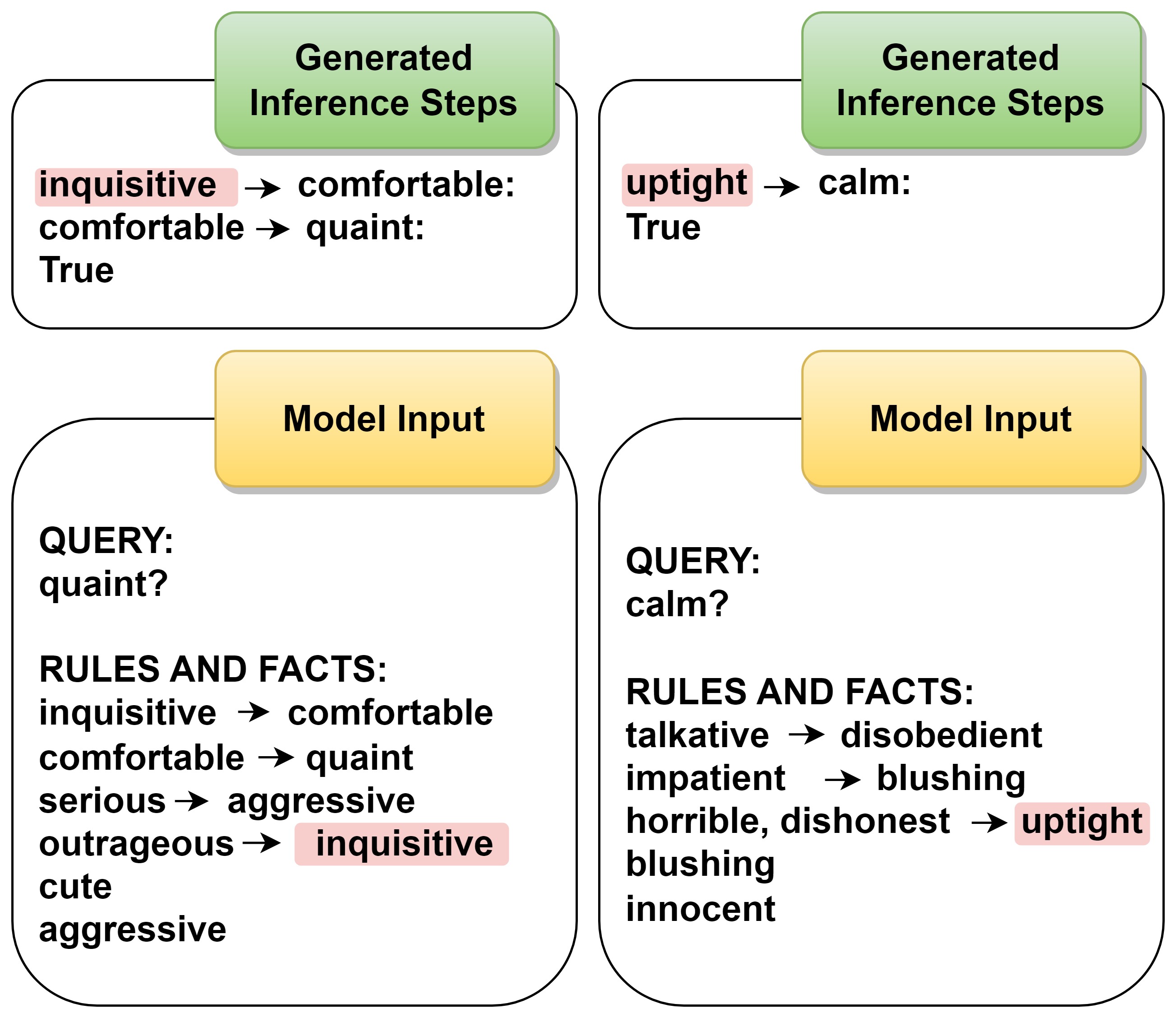}
    \caption{Two examples of where the generated inference steps include inapplicable rules. In both examples the conclusion of the last rule has been mistaken as a fact. The model confuses the conclusion of the last rule in the input string, regarding it as a fact instead of a rule that is not yet satisfied. 
 }
    \label{fig:inapplicable_rule}
\end{figure}

 Inapplicable Rule errors occurred in up to 0.02 percent of the attempted proofs. It was the second most common error for SIP-BART(RP) and SIP-BART(RP\_b) but did not occur at all for SIP-BART(LP), see Table \ref{tab:ConsistencyErrors}. Qualitative analysis suggests that the cause for this error type is the order and location of rules and facts in the input: The input is structured so that the query, rules and facts are located following each other in the input string, and the analysis indicates that the model has mistaken parts of the rules close to the facts, as facts. Two examples of this are shown in Figure \ref{fig:inapplicable_rule}.   

\subsubsection{Spurious Match}
\begin{figure}[htbp]
    \centering
    \includegraphics[width=8.5cm]{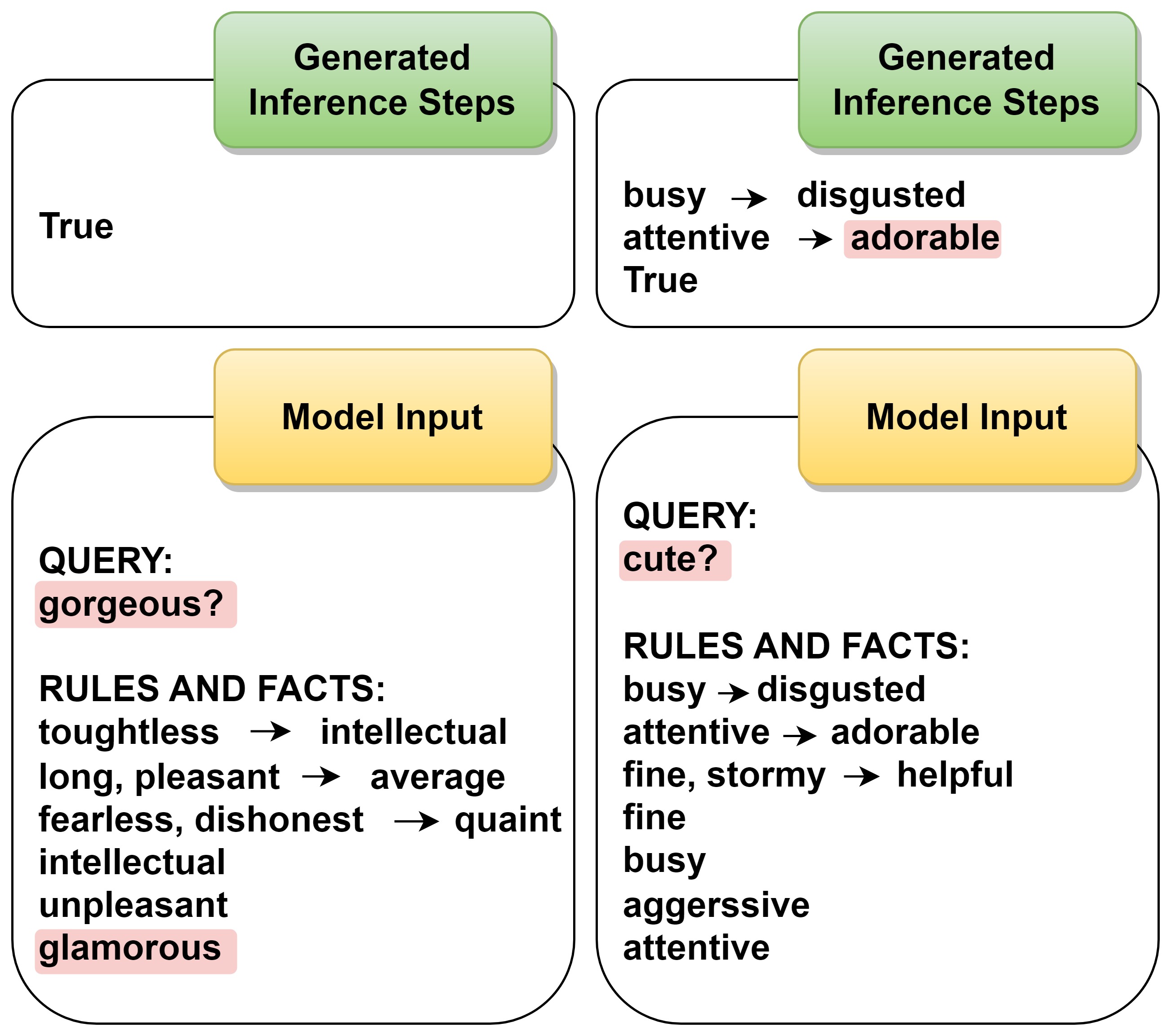}
    \caption{Two examples of where a Spurious Match has occurred. In the first example (left) the fact \emph{glamorous} has been mistaken as the query \emph{gorgeous}. In the second example (right) the conclusion in the generated rule \emph{aggressive,attentive $\Rightarrow$ adorable} is mistaken as the query \emph{cute}.}
    \label{fig:spurious_match}
\end{figure}

 The third type of error, Spurious Match, is the least common error in total, with a prevalence of 0.000 to 0.007 percent. In all cases, the reason was that the model mistakes a synonym word to the query as the actual query and the model predicts the query as True (similar to what happened for Non-existent Rule). In essence, this error means that the system has proved the wrong query. This is exemplified in Figure \ref{fig:spurious_match} where the series of inference steps ends with a rule with the conclusion \emph{adorable} and a True label is generated, but the query to the problem asked for is in fact the synonym \emph{cute}.

\subsubsection{Unexhausted Search Space}
 The final type of error, Unexhausted Search Space, is only applicable for problems predicted to be False, with a prevalence of up to 0.02\%. This was the second most common error for SIP-BART(LP) when tested on RP and RP\_b and was the only error that was found for the false negative problems. 
 No obvious patterns could be observed in why this error occurred, other than that in most cases the model had found all but one rule, which meant that most of these errors occurred on true negatives. This is likely due to the probability of generating the token \emph{False} increases as the search space is getting closer to exhaustion. When that happens, the probability for that token might get too high slightly too quickly, i.e., before the search space is exhausted.

\subsubsection{Summary}
 In summary, even if all the models achieve high levels of consistency, they are not immune to producing errors. In total the most frequent error is Non-existing Rule, but the different SIP-BART models differ somewhat in the distribution of error-types.
 For all the models the accuracy is higher than the consistency, some proofs contain erroneous steps, but still end with the correct Boolean label for the query. However, the difference between accuracy and consistency is quite small, which means that almost all correct classifications are consistent - no model has more than 0.7\% consistency errors on any of the given testing set.
 

 \section{Discussion and Conclusion}

By generating inference steps, SIP-BART minimises the effect of known spurious correlations present in the SimpleLogic dataset. The model divides each problem into smaller sub-problems that are easier to solve, which helps approximate reasoning without exploiting statistical features as shortcuts. This also makes it easier for the model to approach the parts of the problem in the correct order. SIP-BART does this by playing to the strengths of each of its modules. The pattern recognition task, which is handled by the neural-module, is in essence an inductive task where conclusions are drawn from observations. The symbolic module instead handles the deductive part, the transition between each reasoning step. By limiting the work that needs to be done by the neural-module, the SIP-BART model can focus on only trying to find rules that are applicable without the need to take into consideration the previous or the next steps of the inference procedure. 
One explanation for this is the importance of locality in the training data \cite{prystawski2023think}: training on individual proof-steps instead of the final outcome increase locality between the current goal and the desired output (the next applicable rule).

\subsection{Mitigation Strategies} 
Although SIP-BART perform much better than the BERT classifier, it still makes some spurious errors which we identified in our experiments. Below, we also suggest some strategies for mitigating these. Evaluating the strategies fully is left as further work, and should ideally be done on a dataset where these errors are more common than in SimpleLogic to ensure effects are significant.

\subsubsection{Dealing with Synonyms}
All the SIP-BART models were fine-tuned from a pre-trained BART-model. A adverse effect of this became apparent from the quantitative analysis: words were occasionally swapped for synonyms. This makes sense in natural language, but not in logic inferences. All Spurious Matches and many of the Non-existing Rule-errors are due to such synonym-issues. One solution is to train a transformer from scratch, on a restricted vocabulary, instead of starting from a pre-trained model. However, if we want a model that can both reason and understand instructions expressed in natural language, this is not the best solution. This may also require additional computational resources for training.  
Another potential solution could be to use a constrained decoder \cite{geng2023flexible}, which only allows generation of strings adhering to a specified grammar, i.e. only well-formed rules over a specific vocabulary. 

\subsubsection{Dealing with Locality Bias}
The input string describing the problem contains the list of rules, followed by the list of facts.
Inapplicable Rule and Non-existing Rule errors are closely tied to the locality structure of the input and can be considered a spurious statistical feature learned by the models. 
In the Inapplicable Rule-case, 
the models mistakenly treat the tokens toward the end of the input string as facts, instead of the conclusion of a rule, despite facts being demarcated by a different symbol\footnote{e.g. if \emph{aggressive} and \emph{inquisitive} are conclusions to rules, they are written as \emph{aggressive:} and \emph{inquisitive:} in the input, while if they were facts they would be followed by a different symbol, i.e. \emph{aggressive1} and \emph{inquisitive1}.}. 
Similarly, when the model disregards a premise of a rule when making a Non-existing Rule error, it seems like it has not completely internalised the scope of the rules premises, despite that all the literals in the premise are demarcated with a ","-sign. 
A possible way of solving this would be to shuffle the order of the query, rules and facts in the input string of the training data. This might teach the model to avoid focusing on the token's position in the input string when it instead should be paying attention to the demarcation symbols that follow the token.

\subsubsection{Dealing with Search Space Issues}
This strategy relies on a neuro-symbolic architecture (unlike the previous two). A standard approach is to let the neural model generate multiple answers for the same input, then letting the symbolic module check which ones are in fact valid inferences. This will of course induce a small overhead in runtime.
This strategy is particularly relevant for Unexhausted Search Space errors, where we need to compensate for the neural model jumping to conclusions too quickly, before the final inferences has been generated. 

\subsection{Conclusion}
Our study provides empirical evidence as to why inducing transformers to reasoning step by step generally better approximated logical reasoning - it helps avoiding spurious statistical patterns such as correlations between the number of rules and facts and the truth value of the query. Training on whole proofs was not sufficient, the transformer still picked up reasoning shortcuts. Part of the reason for this is likely that training on proof-steps gave more training data: the model would benefit for each step in each training proof. It is also possible that the representation of "failed" proofs for false queries was difficult to learn for WP-BART. However, for computational resource reasons, this was not investigated further.

We also identify four additional types of spurious errors made by the transformer model, arising from replacement of a word by a synonym (an effect of pre-training on natural language), mistaking part of a rule for a fact when located closely in the input or jumping to a final conclusion before completing the proof. We proposed some mitigation strategies to counter this, some of which can be implemented directly for the transformer, but for robust reasoning, we believe it is necessary to implement a neuro-symbolic architecture. Here, a neural module suggests (perhaps multiple) rules and a symbolic module checks them and deals with updates to the problem representation. Another approach is to use a language model just as a semantic parser and train it to translate the whole problem into the input language of a symbolic prover \cite{olausson-etal-2023-linc}. However, one still needs to deal with similar errors as those discussed here causing subtle mistakes in translation to propagate through to the reasoning engine.

\subsection*{Acknowledgement} 
 The computations and storage of data were enabled by resources provided by the National Academic Infrastructure for Supercomputing in Sweden (NAISS) at Chalmers Centre for Computational Science and Engineering (C3SE), partially funded by the Swedish Research Council through grant agreement no. 2022-06725. We would also like to thank Nicholas Smallbone for feedback on a draft of this paper.

\bibliographystyle{splncs04}
\bibliography{refs}

\appendix
\begin{landscape} 
\section{Example SimpleLogic Problem}
        \label{AppA}
        
        \begin{figure}
            \centering
    \includegraphics[scale=0.35]{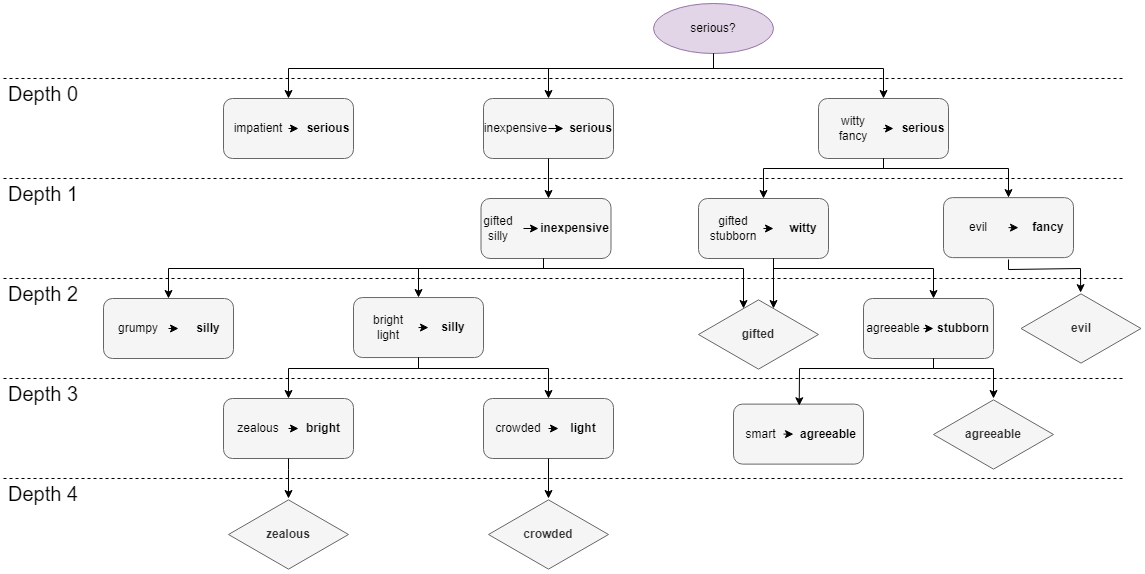}
            \caption{Example of the structure of a problem in SimpleLogic. The query can be seen in the purple circle. The rules are represented in each rectangle, with the premises on the left side and the conclusion in bold text on the right side. The facts are represented as single words in each diamond shape.}
            \label{fig:example}
        \end{figure}
\end{landscape} 

  \section{Hyperparameters and Hardware}
  \label{AppB}
  \setcounter{table}{0}

  We fine-tuned a BART model from Huggingface (\url{https://huggingface.co/facebook/bart-base}) with the default hyperparameters with the exception of batch size and gradient accumulation steps, which was tuned to maximize the memory usage of the GPUs used. For all experiments we used four NVIDIA Tesla A40 GPUs with 48GB RAM.

     \begin{table}[ht]
    \begin{tabular}{|p{50mm}|p{25mm}|}
    \hline
    \textbf{Hyperparameter} & \textbf{Value}  \\ \hline
    Learning rate  & 0.00002\\ 
    Batch size & 8 \\
    Gradient accumulation steps & 4\\
    fp16 & True \\
    Warmup steps & 200 \\
    Weight decay & 0.01 \\
    Epoch & 1 \\
    \hline
    \end{tabular}  
    \centering
 \end{table}

\section{Evaluation of WP-BART}
  \label{AppC}
  \setcounter{table}{0}

  The evaluation based on the precision, recall and F1-score of the worst performing model, WP-BART(LP), on the RP dataset. Together with the precision, recall and F1-score for each of the WP-BART models based on which training and testing datasets where used.

  \begin{table*}[htbp]
    \begin{tabular}{|*{10}{p{14mm}|}}
    \hline
            \textbf{Depth} & \textbf{TPR} & \textbf{FPR} & \textbf{TNR} & \textbf{FNR} & \textbf{Preci-sion} & \textbf{Recall} & \textbf{F1-Score} \\ \hline
        0 & 0.80 & 0. & 0.21 & 0. & 1. & 1. & 1. \\ \hline
        1 & 0.75 & 0. & 0.25 & 0. & 1. & 1. & 1. \\ \hline
        2 & 0.48 & 0. & 0.35 & 0.17 & 1. & 0.74 & 0.85 \\ \hline
        3 & 0.19 & 0. & 0.46 & 0.35 & 1. & 0.35 & 0.52 \\ \hline
        4 & 0.07 & 0.00 & 0.60 & 0.33 & 0.99 & 0.18 & 0.31 \\ \hline
        5 & 0.02 & 0. & 0.70 & 0.28 & 1. & 0.07 & 0.13 \\ \hline
        6 & 0.01 & 0.00 & 0.76 & 0.23 & 0.86 & 0.03 & 0.05 \\ \hline
        \textbf{Total} & \textbf{0.33} & \textbf{0.00} & \textbf{0.47} & \textbf{0.19} & \textbf{0.99} & \textbf{0.63} & \textbf{0.77} \\ \hline
    \end{tabular}
    \centering
    \caption{Evaluation metrics for the WP-BART(LP) tested on RP dataset. In the heading the acronyms for True Positive Rate (TPR), False Positive Rate (FPR), True Negative Rate (TNR) and False Negative Rate (FNR) are used. The 0.00 values are close to zero but not exactly zero. The 0. values are exactly zero. The corresponding information for exactly one is true for 0.99 and 1., respectively.}
    \label{tab:conf_mat_whole_LP_RP}
\end{table*}

\begin{table*}[htbp]
    \begin{tabular}{|*{10}{p{12.5mm}|}}
    \hline
           \textbf{Train} & \textbf{Test} & \textbf{TPR} & \textbf{FPR} & \textbf{TNR} & \textbf{FNR} & \textbf{Preci-sion} & \textbf{Recall} & \textbf{F1-Score} \\ \hline
    LP & LP & 0.431 & 0.001 & 0.507 & 0.061 & 0.999 & 0.877 & 0.934\\ \hline
    LP & RP & 0.332 & 0.000 & 0.474 & 0.194 & 0.999 & 0.631 & 0.773\\ \hline
    LP & RP\_b & 0.346 & 0.000 & 0.474 & 0.180 & 0.999 & 0.657 & 0.793\\ \hline
    RP & LP & 0.462 & 0.218 & 0.290 & 0.030 & 0.680 & 0.940 & 0.789\\ \hline
    RP & RP & 0.487 & 0.117 & 0.358 & 0.038 & 0.807 & 0.927 & 0.863\\ \hline
    RP & RP\_b & 0.492 & 0.113 & 0.362 & 0.034 & 0.814 & 0.936 & 0.871\\ \hline
    RP\_b & LP & 0.461 & 0.137 & 0.371 & 0.031 & 0.771 & 0.937 & 0.846\\ \hline
    RP\_b & RP & 0.482 & 0.069 & 0.405 & 0.044 & 0.874 & 0.917 & 0.895\\ \hline
    RP\_b & RP\_b & 0.488 & 0.065 & 0.410 & 0.037 & 0.883 & 0.929 & 0.905\\ \hline

    \end{tabular}
    \centering
    \caption{Precision, Recall and F1 score for all WP-BART models. In the heading the acronyms for True Positive Rate (TPR), False Positive Rate (FPR), True Negative Rate (TNR) and False Negative Rate (FNR) are used. All numbers are rounded to three decimals. The 0.000 values are close to zero but not exactly zero.}
    \label{tab:conf_mat_whole_all}
\end{table*}

\newpage

\section{Evaluation of SIP-BART}
  \label{AppD}
  \setcounter{table}{0}
  
The evaluation based on the precision, recall and F1-score of the worst performing model, SIP-BART(LP), on the RP dataset. Together with the precision, recall and F1-score for each of the SIP-BART models based on which training and testing datasets where used.

\begin{table*}[htbp]
    \begin{tabular}{|*{10}{p{14mm}|}}
    \hline
    \textbf{Depth} & \textbf{TPR} & \textbf{FPR} & \textbf{TNR} & \textbf{FNR} & \textbf{Preci-sion} & \textbf{Recall} & \textbf{F1-Score} \\ \hline
    0 & 0.795 & 0.000 & 0.204 & 0. & 0.999 & 1. & 0.999 \\ \hline
    1 & 0.754 & 0.000 & 0.245 & 0. & 0.999 & 1. & 0.999 \\ \hline
    2 & 0.652 & 0. & 0.347 & 0. & 1. & 1. & 1. \\ \hline
    3 & 0.537 & 0.000 & 0.461 & 0.000 & 0.999 & 0.999 & 0.999\\ \hline
    4 & 0.400 & 0.001 & 0.598 & 0.000 & 0.998 & 0.999 & 0.998\\ \hline
    5 & 0.297 & 0.001 & 0.701 & 0.000 & 0.997 & 0.998 & 0.997\\ \hline
    6 & 0.239 & 0.005 & 0.755 & 0.000 & 0.980 & 0.998 & 0.989\\ \hline
    \end{tabular}
    \centering
    \caption{Evaluation metrics for the SIP-BART(LP) tested on RP dataset. In the heading the acronyms for True Positive Rate (TPR), False Positive Rate (FPR), True Negative Rate (TNR) and False Negative Rate (FNR) are used. The values 0.000 are close to zero but not exactly zero. The values that are 0. are exactly zero, an example is found for depth 2 for FP, and means that not a single case was found. The values 1. means exactly 1.}
    \label{tab:conf_mat_step_LP_RP}
\end{table*}

\begin{table*}[htbp]
    \begin{tabular}{|*{10}{p{12.5mm}|}}
    \hline
       \textbf{Train} & \textbf{Test} & \textbf{TPR} & \textbf{FPR} & \textbf{TNR} & \textbf{FNR} & \textbf{Preci-sion} & \textbf{Recall} & \textbf{F1-Score} \\ \hline
    LP & LP & 0.492 & 0.000 & 0.507 & 0.000 & 0.999 & 0.999 & 0.999\\ \hline
    LP & RP & 0.525 & 0.001 & 0.473 & 0.000 & 0.998 & 0.999 & 0.998\\ \hline
    LP & RP\_b & 0.525 & 0.002 & 0.472 & 0.000 & 0.997 & 0.999 & 0.998\\ \hline
    RP & LP & 0.492 & 0.000 & 0.507 & 0. & 0.999 & 1. & 0.999\\ \hline
    RP & RP & 0.525 & 0. & 0.474 & 0.000 & 1. & 0.999 & 0.999\\ \hline
    RP & RP\_b & 0.525 & 0. & 0.475 & 0. & 1. & 1. & 1.\\ \hline
    RP\_b & LP & 0.492 & 0.000 & 0.507 & 0.000 & 0.999 & 0.999 & 0.999\\ \hline
    RP\_b & RP & 0.525 & 0.000 & 0.474 & 0.000 & 0.999 & 0.999 & 0.999\\ \hline
    RP\_b & RP\_b & 0.525 & 0. & 0.474 & 0.000 & 1. & 0.999 & 0.999\\ \hline

    \end{tabular}
    \centering
    \caption{Precision, Recall and F1 score for all WP-BART models. In the heading the acronyms for True Positive Rate (TPR), False Positive Rate (FPR), True Negative Rate (TNR) and False Negative Rate (FNR) are used. The 1. values are exactly one. All other numbers are rounded to three decimals, except those very close to 1.  Values very close to 1 have been floored to 0.999. Similarly, the 0.000 values are close to zero but not exactly zero. The 0. values are exactly zero, an example is row 5 for FP, and means that not a single case was found.}
    \label{tab:conf_mat_step_all}
\end{table*}

\end{document}